\documentclass[journal]{IEEEtran}
\usepackage{cite}
\usepackage{hyperref}
\usepackage{multirow} %
\usepackage{longtable} 
\usepackage{array} 
\usepackage{makecell}%
\usepackage{threeparttable} 
\usepackage{booktabs}

\usepackage{graphicx}
\ifCLASSINFOpdf
\else
\fi
\usepackage{amsmath}
\usepackage{algorithmic}
\usepackage{algorithm}

\usepackage{array}
\usepackage{color}
\usepackage{cleveref}
\usepackage{CJK}
\usepackage{array}
\usepackage{booktabs}
\begin{document}
%
\title{Global and Local Contrastive Self-Supervised Learning for Semantic Segmentation of HR Remote Sensing Images}
%
%
%
\author{
    Haifeng~Li, 
    Yi Li, 
    Guo Zhang, 
    Ruoyun Liu, 
    Haozhe Huang, 
    Qing Zhu, 
    Chao Tao
   
\thanks{H. Li, Y. Li, R. Liu, H. Huang and C. Tao are with the School of Geosciences and Info-Physics, Central South University, Changsha, 410083, China. Email: lihaifeng@csu.edu.cn. C. Tao is the corresponding author. Email: kingtaochao@csu.edu.cn.}
\thanks{G. Zhang is with the State Key Laboratory of Information Engineering in Surveying, Mapping and Remote Sensing, Wuhan University, Wuhan 430079, China.}
\thanks{Q. Zhu is with the Faculty of Geosciences and Environmental Engineering, Southwest Jiaotong University, Chengdu 611756, China.}
\thanks{Citation: Haifeng Li, Yi Li, Guo Zhang, Ruoyun Liu, Haozhe Huang, Qing Zhu, Chao Tao. Global and Local Contrastive Self-Supervised Learning for Semantic Segmentation of HR Remote Sensing Images. IEEE Transactions on Geoscience and Remote Sensing. 2022. 10.1109/TGRS.2022.3147513.}
    }

\maketitle

\begin{abstract}
Recently, supervised deep learning has achieved great success in remote sensing image (RSI) semantic segmentation. However, supervised learning for semantic segmentation requires a large number of labeled samples, which is difficult to obtain in the field of remote sensing. A new learning paradigm, self-supervised learning (SSL), can be used to solve such problems by pre-training a general model with a large number of unlabeled images and then fine-tuning it on a downstream task with very few labeled samples. Contrastive learning is a typical method of SSL that can learn general invariant features. However, most existing contrastive learning methods are designed for classification tasks to obtain an image-level representation, which may be suboptimal for semantic segmentation tasks requiring pixel-level discrimination. Therefore, we propose a global style and local matching contrastive learning network (GLCNet) for remote sensing image semantic segmentation. Specifically, 1) the global style contrastive learning module is used to better learn an image-level representation, as we consider that style features can better represent the overall image features. 2) The local features matching contrastive learning module is designed to learn representations of local regions, which is beneficial for semantic segmentation. We evaluate four RSI semantic segmentation datasets, and the experimental results show that our method mostly outperforms state-of-the-art self-supervised methods and the ImageNet pre-training method. Specifically, with 1\% annotation from the original dataset, our approach improves Kappa by 6\% on the ISPRS Potsdam dataset relative to the existing baseline. Moreover, our method outperforms supervised learning methods when there are some differences between the datasets of upstream tasks and downstream tasks. Our study promotes the development of self-supervised learning in the field of RSI semantic segmentation. Since SSL could directly learn the essential characteristics of data from unlabeled data, which is easy to obtain in the remote sensing field, this may be of great significance for tasks such as global mapping. The source code is available at https://github.com/GeoX-Lab/G-RSIM.
\end{abstract} 


\IEEEpeerreviewmaketitle

\section{Introduction}

\IEEEPARstart{W}{ith} the development of remote sensing techniques, high-resolution (HR) satellite images are easily obtained. Remote sensing images are widely used in urban planning, disaster monitoring, environmental protection, agricultural management, etc. \cite{SchumannBrakenridge-33,WeissJacob-application,ShiZhong-85,ChenYuan-91,ZhuLiao-92}. The extraction and recognition of information from remote sensing images is the basis of these applications. Semantic segmentation, as a pixel-level image analysis technology, is one of the most important and challenging research directions in the remote sensing image interpretation field \cite{XingSieber-34}.

Traditional RSI semantic segmentation algorithms are mostly machine learning approaches based on handcrafted features, such as support vector machines (SVMs, \cite{Cortes1995-svm,PalMather-35}), random forests (RFs, \cite{Breiman2001-RF}), and artificial neural networks (ANNs, \cite{Basse2014-ANN}). Since AlexNet \cite{KrizhevskySutskever-23} won the ILSVR champion in 2012, deep learning, especially deep convolution neural networks (DCNNs), has attracted increasing attention \cite{SimonyanZisserman-24,SzegedyLiu-36,He2016-25}. Compared with traditional methods, deep learning is completely data-driven and can extract more abstract high-level features, achieving remarkable results on image classification tasks \cite{SzegedyVanhoucke-37}. Subsequently, full convolution network-based approaches, such as FCN \cite{Long2015-26}, U-Net \cite{Ronneber2015-27}, and the DeepLab series \cite{ChenPapandreou-28deeplabv2,ChenPapandreou-30-deeplabv3,ChenZhu-31-deeplabv3p}, almost dominate the field of computer vision image semantic segmentation. In remote sensing, researchers have improved the general semantic segmentation network, taking into account specific characteristics of remote sensing and further improving the accuracy of RSI semantic segmentation tasks \cite{WaldnerDiakogiannis-69,LiuMi-84}. For example, Mohammadimanesh et al. \cite{MohammadimaneshSalehi-59-PolSAR} designed a new FCN architecture specifically for the classification of wetland complexes using polarimetric synthetic aperture radar (PolSAR). Ding et al. \cite{DingBruzzone-60} focused on the problem of remote sensing image semantic segmentation with a large image size. To better exploit global context information in remote sensing images, they propose a two-stage semantic segmentation network, the first stage of which is scaling the image to different sizes to obtain global contextual information and local detail information and then fusing the features to improve the accuracy.
\par
However, deep learning-based supervised RSI semantic segmentation methods rely heavily on a large number of high-quality labeled samples. As RSI semantic segmentation technology plays an increasingly important role in global sustainable development, its need for a large number of global-wise, high-quality labeled samples is growing \cite{YuanShen-75,ValiComai-76}. Semantic segmentation requires pixel-level labeling, which is very costly. Furthermore, because remote sensing images vary greatly in time and space, the existing labeled data are only interceptions of the images; it is difficult to obtain a large number of annotated samples with extremely high richness that cover global areas, multiple resolutions, season and spectra. To solve the problem of an insufficient number of labeled samples, one strategy is to generate more samples through data augmentation \cite{ShortenKhoshgoftaar-47-dataaug}, generative adversarial networks (GAN) \cite{BowlesChen-48-GAN-aug}, etc.; a second strategy is to use other annotated data, such as pre-training \cite{Frid-AdarBen-Cohen-49} or transfer learning \cite{CuiChen-50-trans,GengDeng-83}, which aims to transfer the knowledge learned from a larger or more related domain to improve performance on the target domain or reduce the dependence on labeled samples; another strategy is to learn how to have better performance on only a few labeled samples, such as meta-learning \cite{LiCui-46-RS-MetaNet}. However, all the above methods are based on the paradigm of supervised learning, which is highly related to specific tasks and datasets, and it is impossible to obtain a general model. For example, transfer learning may have negative transfer when the difference between the source domain and the target domain is large \cite{VuJain-56}. In addition, these methods do not make use of the abundant unlabeled data.
\par
\begin{figure}[!ht]
    \centering
    \includegraphics[width=3in]{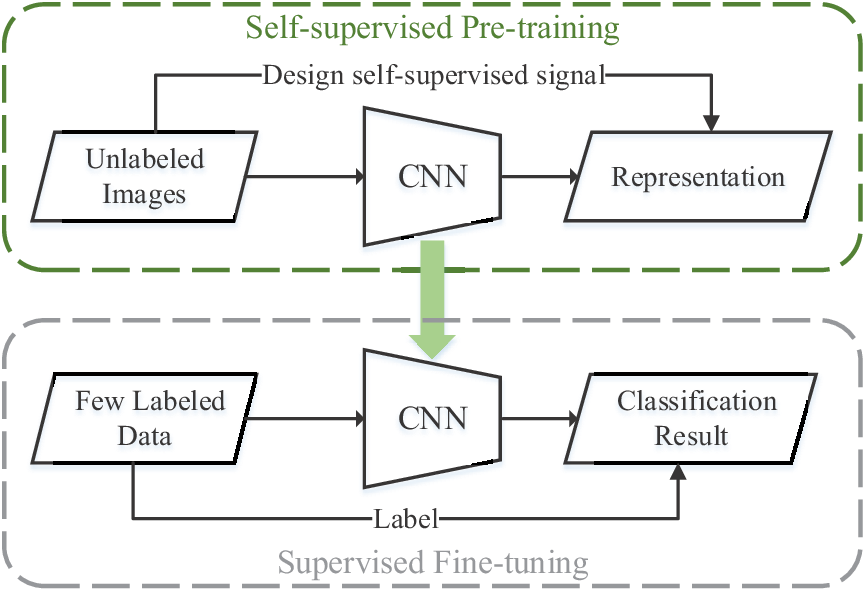}
    \caption{The self-supervised learning paradigm}
	\label{fig1:SSL}
\end{figure}

Self-supervised learning\cite{jing2020self,misra2020self,jaiswal2021survey} provides a new paradigm, as shown in Fig. \ref{fig1:SSL}, which first learns knowledge from unlabeled image data by designing self-supervised signals and then transfers it to downstream tasks, achieving comparable performance to supervised learning on downstream tasks with a limited number of labeled samples \cite{jing2020self}. While large amounts of labeled data are not available for remote sensing images, unlabeled image data covering the whole world with great diversity and richness are readily accessible, and because the information contained in the unlabeled image data is much richer than the sparse labels, we can expect to potentially learn more general knowledge through self-supervised learning.
\par
In this work, we focus on contrastive learning, which is a typical and successful self-supervised method in the field of natural image classification \cite{HeFan-42-MoCo,Chen2020-simCLR,LiZhou-45-PCL}. Recently, some studies have introduced contrastive learning into the field of remote sensing and proved its feasibility \cite{StojnicRisojevic-87,LiuYu-88,TaoQi-89,JungOh-90}. However, most of these studies are designed for scene classification tasks. It can be expected that it is appropriate to utilize existing instance-wise contrastive learning \cite{Chen2020-simCLR} on scene classification datasets, but, as producing scene classification datasets is complex, it is more meaningful to explore how to perform contrastive learning directly on cropped remote sensing images. In addition, the existing instance-wise contrastive learning is designed for learning image-level representations, thus it is not optimal for semantic segmentation tasks that require pixel-level discrimination.
\par
For these reasons, in this work, we introduce the contrastive learning paradigm into RSI semantic segmentation. In the pre-training stage, we use contrastive learning to enhance the consistency of the sample on the label-free data to learn a General Remote Sensing vIsion model(G-RSIM). G-RSIM enhances invariance, such as illumination invariance, rotation invariance, and scale invariance. Second, the existing instance-wise contrastive learning is mainly designed for image classification tasks and only focuses on the learning of global representations. However, there is a balance between global feature learning and local feature learning for RSI semantic segmentation tasks: from the perspective of global representations, remote sensing images have overall differences due to disparities in time (spring, summer, autumn, and winter), weather, sensors, etc.; from the perspective of local representations, pixel-level semantic segmentation requires more local information \cite{LANet-85}. Therefore, we propose the global style and local matching contrastive learning network (GLCNet) framework, in which the global style contrastive learning module focuses on the global representation, and the local matching contrastive learning module is used to learn pixel (local)-level features.
\par
The main contributions of this paper are summarized as follows:

1) We apply self-supervised contrastive learning to remote sensing image semantic segmentation tasks and verify it on multiple datasets. The model can directly learn features from unlabeled images to guide the downstream semantic segmentation tasks with limited annotations;

2) We propose a new self-supervised contrastive learning framework, namely the global style and local matching contrastive learning network (GLCNet), which focuses on balancing the global and local feature learning in remote sensing image semantic segmentation tasks;

3) We evaluate our proposed method on two public datasets and two realistic datasets. Experimental results show that our method outperforms other self-supervised methods. It also outperforms the supervised learning method in the situation where the upstream and downstream datasets are not highly similar.
\par
The remainder of this paper is organized as follows. In Section II, we introduce the semantic segmentation approach based on self-supervised contrastive learning and our further improved self-supervised approach for the semantic segmentation task – GLCNet. The experiments and results are provided in Section III. Section IV states further discussion, and Section V presents the conclusion.
\begin{figure*}[!ht]

    \centering
    \includegraphics[width=6in]{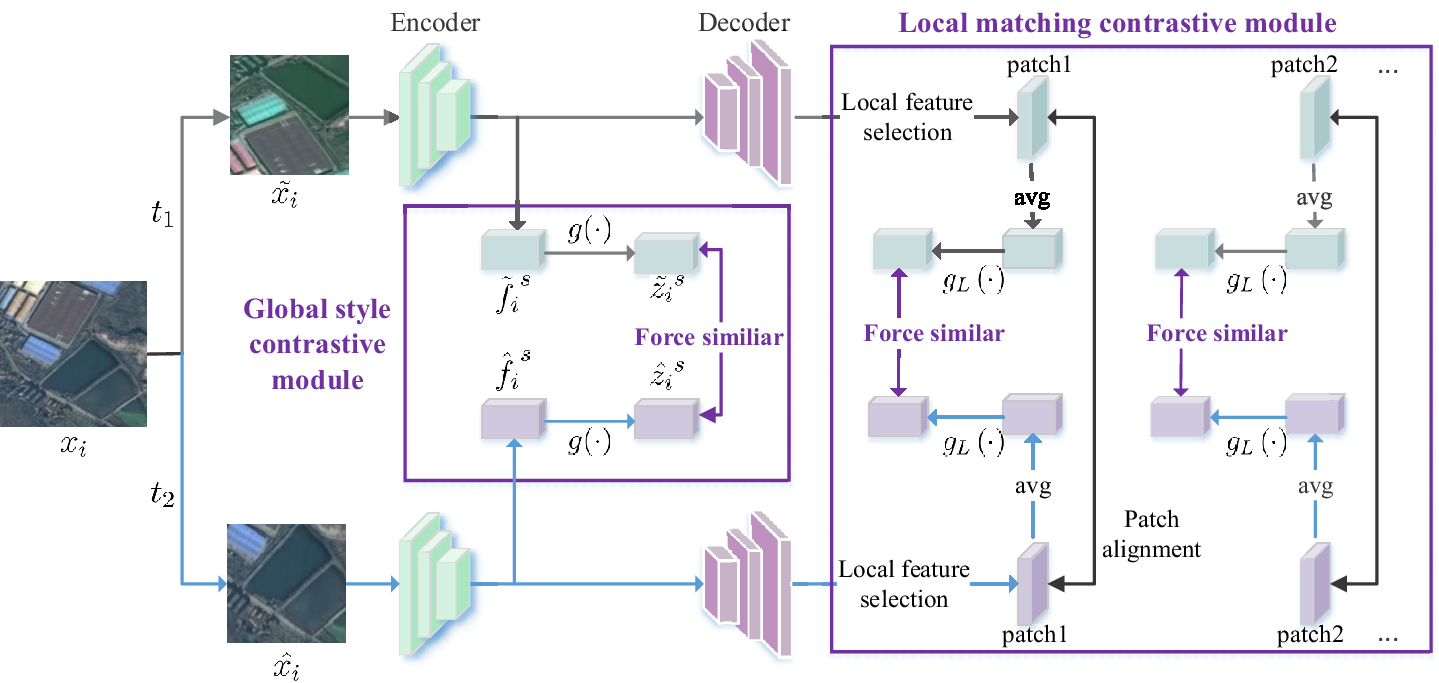}
    \caption{Schematic illustration of our proposed GLCNet architecture for the self-supervised pretext task. (The figure omits the display of negative samples. For the global style contrastive learning module, negative samples are other samples from a mini-batch. For the local matching contrastive learning module, all other local regions are negative samples except for the matching local regions.)}

	\label{fig:GLCNet}
\end{figure*}
\section{Method}
\subsection{Overview}
The success of supervised deep learning relies on a large number of labeled samples, which is difficult to achieve in RSI semantic segmentation. As shown in Fig. \ref{fig1:SSL}, self-supervised learning provides a new paradigm for learning potentially useful knowledge directly from a large amount of readily available unlabeled data and then transferring it to downstream tasks to achieve better performance, especially with limited labeled samples. In our work, the downstream task is the semantic segmentation of remote sensing images, as such, we concentrate on designing a self-supervised model for the semantic segmentation of remote sensing images. In this paper, we introduce contrastive learning to learn the general invariant representation. Simultaneously, we propose the GLCNet self-supervised method for considering the characteristics of the semantic segmentation task, as shown in Fig. \ref{fig:GLCNet}. The GLCNet self-supervised method mainly contains two modules:
\par
1) The global style contrastive learning module mainly considers that the feature generated by global average pooling used in the existing contrastive learning is not a good substitute for the overall feature of an image. Thus, the style features that are more representative of the overall features of an image are introduced to help the model better learn global representations;
\par

2) The local matching contrastive learning module is proposed mainly for the following two reasons. First, the land cover categories in a single image in the semantic segmentation dataset are extremely rich. Extracting only the global features of the whole image to measure and distinguish images will lose much useful information. Second, the image-level representations obtained by instance-wise contrastive learning may be sub-optimal for semantic segmentation tasks that require pixel-level discrimination.

\subsection{Contrastive learning}
\begin{figure}[!ht]
    \vspace{0.15cm}
    \setlength{\abovecaptionskip}{0.1cm}
    \setlength{\belowcaptionskip}{0.15cm} 
    \centering
    \includegraphics[width=2.5in]{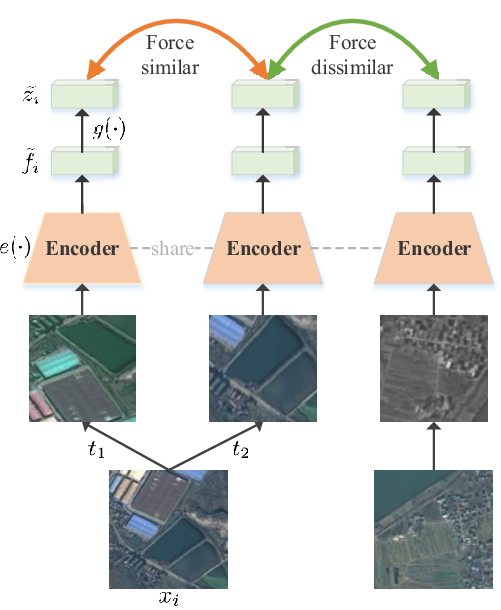}
    \caption{Illustration of contrastive learning pretext task}

	\label{fig:CL}
\end{figure}

Contrastive learning learns by forcing positive sample pairs to be similar and negative sample pairs to be dissimilar \cite{WuXiong-52-instance-discrimination,Chen2020-simCLR}. The key to contrastive learning methods is to construct positive and negative samples. The most recent breakthrough methods \cite{HeFan-42-MoCo,Chen2020-simCLR} classify instances as their own labels, which means that different enhanced versions of a sample are treated as positive samples, and other samples are treated as negative samples. Contrastive learning encourages the model to learn the invariance of transformations and the ability to distinguish different samples. In this work, we use contrastive learning to learn general spatiotemporal invariance features for remote sensing. Specifically, we perform random rotation, cropping, scaling and other data augmentation operations on samples to make the model learn spatial invariances such as rotation invariance and scale invariance. In addition, the temporal difference of remote sensing images mainly lies in the overall texture and color differences caused by seasonal factors and imaging conditions. Due to the lack of multitemporal image data, we simulate the time transformation by applying random color distortion, random noise, etc. on the samples to make the model learn temporal-invariant features.
\par
Inspired by SimCLR \cite{Chen2020-simCLR}, we apply contrastive learning to train the encoder of the semantic segmentation network, as shown in Fig. \ref{fig:CL}, which consists of the following four main components:
\par
\subsubsection{Data augmentation}
\begin{figure}[!ht]
   
    \centering
    \includegraphics[width=3in]{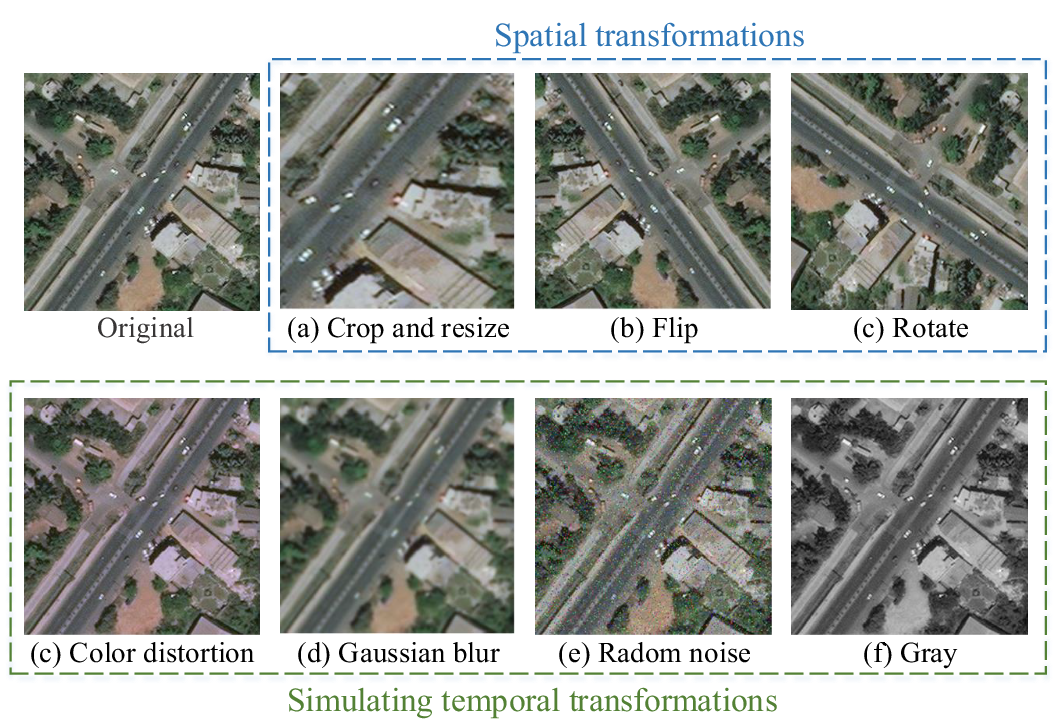}
    \caption{Illustrations of the studied data augmentation operators}
	\label{fig:data_aug}
\end{figure}
To encourage the model to learn general spatiotemporal invariance features, as shown in Fig. \ref{fig:data_aug}, we perform spatial transformations such as random cropping, resizing, flipping, and rotation for the learning of spatial invariance features and simulate temporal transformations with color distortion, Gaussian blur, random noise, etc. for the learning of temporal invariance features. Specifically, two augmented views $\tilde{x}$ and $\hat{x}$ are generated from a given sample $x$ by data augmentation $t_1$ and $t_2$, i.e. $\tilde{x}=t_1(x)$, $\hat{x}=t_2(x)$. In this work, $t_1$ represents random cropping followed by resizing to a fixed resolution (e.g., 224 × 224), and $t_2$ represents sequentially applying several augmentations: random cropping followed by resizing to a fixed resolution, random flipping, random rotating, random color distortion, random Gaussian blur, random noise, and random gray.
\subsubsection{Feature extraction}
Global features are extracted from the augmented sample instances using an encoder network $e(\cdot)$:

\begin{equation}
{\tilde {f_i}=\mu(e(\tilde{x_i})), \hat {f_i}=\mu(e(\hat{x_i}))\label{con:feature extraction} }
\end{equation}

where $\mu$ represents the calculation of the mean value of each channel in the feature map, i.e., the global average pooling. In this work, $e(\cdot)$ is the encoder of the semantic segmentation network DeepLabV3+ \cite{ChenZhu-31-deeplabv3p}.
\subsubsection{Projection head}
As shown in Equation \ref{con:proj_head}, projection head $g(\cdot)$ is an MLP with one hidden layer (with ReLU). The presence of $g(\cdot)$ in SimCLR \cite{Chen2020-simCLR} has been proven to be very beneficial, possibly because it allows the $e(\cdot)$ to form and retain more potentially useful information for downstream tasks.
\begin{equation}
{\tilde {z_i}=g(\tilde{f_i})=W^{(2)}R(W^{(1)}\tilde{f_i}), \hat {z_i}=g(\hat{f_i})\label{con:proj_head} }
\end{equation}
where $R$ is a ReLU nonlinearity.
\subsubsection{Contrastive loss}
Contrastive loss expects positive sample pairs to be similar and negative sample pairs to be dissimilar. Specifically, N samples from a minibatch are augmented to be 2N samples. A pair of samples augmented from the same sample form a positive pair, and the other 2(N-1) samples are negative samples. Thus, the contrastive loss $\mathcal{L}_C$ is defined as
\begin{subequations}\label{con_loss}
\begin{equation}
{\mathcal{L}_C=\frac{1}{2N}\sum_{k=1}^N \left( \ell\left( \tilde{x_i},\hat{x_i} \right)+ \ell\left( \hat{x_i},\tilde{x_i} \right) \right) \label{con:loss} }
\end{equation} 
with:
\begin{equation}
\ell\left( \tilde{x_i},\hat{x_i} \right)=-\log {\frac{\exp{(sim(\tilde {z_i},\hat {z_i})/\tau)}}{\begin{matrix} \sum_{x\in\Lambda^-} \exp{(sim(\tilde{z_i},g(f(x)))/\tau)} \end{matrix}  \label{con:loss_t}}}
\end{equation}
\end{subequations}
where $sim$ denotes the similarity measure function between two feature vectors and, in this work, is the cosine similarity. $\Lambda^-$ denotes $2(N-1)$ negative samples in addition to the positive sample pair, and $\tau$ denotes a temperature parameter.
\par
Although powerful image-level representations can be learned by the existing contrastive learning paradigm, there are still some problems. First, the existing contrastive learning uses global average pooling features for feature extraction of a sample, which may not be a good representation of the overall characteristics of a sample. Second, and more critically, the image-level representation learned by instance-wise contrastive learning may be suboptimal for semantic segmentation tasks that require pixel-level discrimination. Therefore, we proposed the GLCNet.
\subsection{Global style and Local matching Contrastive learning Network (GLCNet)}
Our proposed GLCNet method is shown in Fig. \ref{fig:GLCNet}, and mainly contains two modules: the global style contrastive learning module, which mainly focuses on the problem that the feature generated by global average pooling used in existing contrastive learning is not a good substitute for the overall feature of a complex remote sensing image; and the local matching contrastive learning module, which mainly considers that most existing contrastive learning methods are designed for image classification tasks to obtain image-level features and therefore may be suboptimal for semantic segmentation requiring pixel-level discrimination. The details are as follows.
\subsubsection{Global style contrastive learning module}
 Global style contrastive learning learns by forcing different augmented views of a sample to be similar but dissimilar to other samples, which is analogous to the existing instance-wise contrastive learning method. The difference is that we use style features instead of the simple average pooling features used in instance-wise contrastive learning, as we believe it is more representative of the overall features of an image. Huang and Belongie \cite{HuangBelongie-53-style} indicate that the channel-wise mean and variance of the feature map extracted by a CNN can represent the style features of an image, so we calculate the channel-wise mean and variance of the features extracted by the encoder $e(\cdot)$ to extract the global style feature vector, which is defined as
 
\begin{equation}
{
f_s\left(x_{i} \right)=\operatorname{concat}\left(\mu\left(e\left(x_{i}\right)\right), \sigma\left(e\left(x_{i}\right)\right)\right) \label{style_f}
}
\end{equation}
where $\mu$ denotes the channel-wise mean of the feature map and $\sigma$ denotes the channel-wise variance. 
\par
Therefore, for N samples from a mini-batch, similar to Equation \ref{con_loss}, the global style contrastive learning loss is defined as follows

\begin{subequations}\label{Global_loss}
\begin{equation}
{\mathcal{L}_G=\frac{1}{2N}\sum_{k=1}^N \left( \ell_g\left( \tilde{x_i},\hat{x_i} \right)+ \ell_g\left( \hat{x_i},\tilde{x_i} \right) \right) \label{Global_con_loss} }
\end{equation}
with:

\begin{equation}
{\ell_g\left( \tilde{x_i},\hat{x_i} \right)=-\log {\frac{\exp{(sim(\tilde{{z_i}}^s,\hat{{z_i}}^s)/\tau)}}{\begin{matrix} \sum_{x\in\Lambda^-} \exp{\left(\frac{sim(\tilde{{z_i}}^s,g(f_s(x)))}{\tau}\right)} \end{matrix} \label{Global_con_loss_t}}}}
\end{equation}
\end{subequations}
where, $\tilde{{z_i}}^s=g(f_s(\tilde{x_i}))$,$\hat{{z_i}}^s=g(f_s(\hat{x_i}))$
\subsubsection{Local matching contrastive learning}
The local matching contrastive learning module is proposed mainly for the following two reasons. First, the land cover categories in a single image in the semantic segmentation dataset are extremely rich. Extracting only the global features of the whole image to measure and distinguish images will result in the loss of much information; second, instance-wise contrastive learning methods are used to obtain image-level features that may be suboptimal for semantic segmentation requiring pixel-level discrimination. Therefore, the local matching contrastive learning module is designed to learn the representation of local regions, which is beneficial for pixel-level semantic segmentation. It consists of the following main components:
\par
\begin{figure}[ht]
    \vspace{0.15cm} 
    \setlength{\abovecaptionskip}{0.1cm} 
    \setlength{\belowcaptionskip}{0.15cm} 
    \centering
    \includegraphics[width=3in]{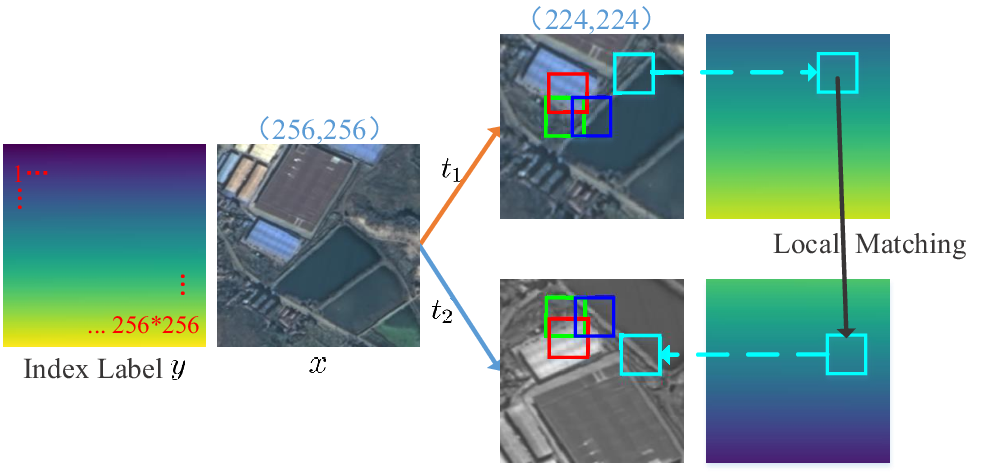}
    \caption{Schematic diagram of local area selection and matching}

	\label{fig:local_matching}
\end{figure}
\textit{a). Local region selection and matching}
\par
As shown in Fig. \ref{fig:local_matching}, two transformed versions $\tilde{x}$ and $\hat{x}$ are from the same image $x$, i.e. $\tilde{x}=t_1\left(x\right), \hat{x}=t_2\left(x\right)$, we select and match multiple local regions from $\tilde{x}$ and $\hat{x}$. Data augmentation operations such as random cropping, flipping, and rotating result in the mismatching positions between $\tilde{x}$ and $\hat{x}$. Therefore, we record the pixel position by introducing an index label to ensure that the center positions of the two matching local regions correspond to each other in the original image. Specifically, we first randomly select a local region with a size of $s_p\times s_p$ from $\tilde{x}$, and then the location of the matching local region with the same size in $\hat{x}$ is determined according to the index value of the center position of the local region in $\tilde{x}$. In addition, to ensure that there is no excessive overlap between different local regions, the local area is excluded after each selection, guaranteeing that the center of the subsequently selected local region does not fall into the previously selected local regions. The above steps are repeated several times to obtain multiple matching local regions.
\par 
\textit{b). Local matching feature extraction}
\par
The local feature extraction steps are as follows: First, the feature maps $d\left(e\left(\tilde{x}\right)\right)$ and $d\left(e\left(\hat{x}\right)\right)$ are extracted from a pair of positive samples $\left(\tilde{x},\hat{x}\right)$ from the encoder-decoder CNN network. In this work, $e(\cdot)$ and $d(\cdot)$ correspond to the encoder and decoder of DeepLabV3+ \cite{ChenZhu-31-deeplabv3p}, respectively. Second, the local feature maps of multiple matching local regions are obtained from $d\left(e\left(\tilde{x}\right)\right)$ and $d\left(e\left(\hat{x}\right)\right)$ according to the idea of selecting and matching local regions in \textit{A}. If $\tilde{p}_j$ and$\hat{p}_j$ are feature maps of matching local regions, where $\tilde{p}_j$ comes from $d\left(e\left(\tilde{x}\right)\right)$ and $\hat{p}_j$ comes from $d\left(e\left(\hat{x}\right)\right)$, the final local feature vector is defined as follows:
    
\begin{equation}
\tilde{f}_L^j=f_{L}\left(\tilde{p_j}\right)=\mu\left(\tilde{p_j}\right), \hat{f}_L^j=\mu\left(\hat{p_j}\right)
\end{equation}
where $\mu$ represents the calculation of the mean value of each channel in the feature map.
\par 
\textit{c). Local matching contrastive loss}
\par
The local matching contrastive loss updates a complete semantic segmentation encoder-decoder network by forcing the feature representations of matching local regions to be similar and the feature representations of different local regions to be dissimilar. For N samples from a mini-batch, the local matching contrastive loss is defined as follows:

\begin{subequations}\label{local_loss}
\begin{equation}
{\mathcal{L}_L=\frac{1}{2N_{L}} \sum_{j=1}^{N_{L}} \left(\ell_{L}\left( \tilde{p_j}, \hat{p_j}\right)+\ell_{L}\left(\hat{p_j}, \tilde{p_j}\right)\right)}
\end{equation}
with:

\begin{equation}
{\ell_{L}\left( \tilde{p_j}, \hat{p_j}\right)=-\log {\frac{\exp{(sim\left(\tilde{{\mu_j}},\hat{{\mu_j}}\right)/\tau)}}{\begin{matrix} \sum_{p\in\Lambda_L^-} \exp{\left(\frac{sim(\tilde{{\mu_j}},g_L(f_L(p)))}{\tau}\right)} \end{matrix} \label{local_con:loss_t}}}}
\end{equation}
\begin{equation}
\tilde{\mu_j}=g_L(\tilde{f}_L^j)=g_L\left(f_{L}\left(\tilde{p_j}\right)\right), \hat{\mu_j}=g_L(\hat{f}_L^j)
\end{equation}
\end{subequations}
where, $N_L$ denotes the number of all local regions selected from a mini-batch of N samples, i.e. $N_L=N\times n_p$, where $n_p$ is the number of matched local regions obtained from a sample. $\Lambda_L^-$ is a set of feature maps corresponding to all local regions except the two matched local regions, and $g_L(\cdot)$ is a projection head that is similar to $g(\cdot)$.
\subsubsection{Total loss}
Global style contrastive learning can capture global information, and local matching contrastive learning focuses on learning representations of local regions, both of which are important for semantic segmentation tasks. When only global style contrastive learning is available, the learned image-level representations are not optimal for semantic segmentation tasks; when only local matching contrastive learning is available, local regions will be over-distinguished, which tends to lead to features belonging to the same category being pushed further away. Therefore, our approach consists of these two parts, that is, the final loss is defined as follows:

\begin{equation}\label{total_loss}
\mathcal{L}=\lambda \cdot \mathcal{L}_G+(1-\lambda)\mathcal{L}_L
\end{equation}
where, in this work, $\lambda$ is the constant 0.5. $\mathcal{L}_G$ represents the global style contrastive loss in Equation \ref{Global_loss}, which is only used to update the encoder network. And $\mathcal{L}_L$ represents the local matching contrastive loss in Equation \ref{local_loss} , which can simultaneously update both the encoder and decoder.
\begin{algorithm}[!htb] 
	\caption{Algorithm of GLCNet Pre-training} 
	\label{alg:GLCNet}
	\begin{algorithmic}[1] 
	\REQUIRE A set of images $X$; structure of $e(\cdot)$, $d(\cdot)$, $g(\cdot)$, $g_L(\cdot)$; augmentation $t_1$,$t_2$; batch size $N$; parameters $\tau$, $\lambda$, $s_p$, $n_p$
	\FOR {sampled mini-batch $x=\{\boldsymbol{x}_k\}_{k=1}^{N}$ from $X$ }
	    \STATE Build index label $y=\{\boldsymbol{y}_k\}_{k=1}^{N}$ using the size of $x$\\
	    \FORALL{}
	        \STATE Draw augmentations: $\tilde{x_k}=t_1(x_k)$, $\tilde{y_k}=t_1(y_k)$, $\hat{x_k}=t_2(x_k)$, $\hat{y_k}=t_2(y_k)$
	    \ENDFOR
	    \STATE Get global style features: \\$f_s\left(\tilde{x} \right)=\operatorname{concat}\left(\mu\left(e\left(\tilde{x}\right)\right),\sigma\left(e\left(\tilde{x}\right)\right)\right)$,\\ $f_s\left(\hat{x} \right)=\operatorname{concat}\left(\mu\left(e\left(\hat{x}\right)\right),\sigma\left(e\left(\hat{x}\right)\right)\right)$
	    \STATE Get local matching feature maps $\tilde{p}$ and $\hat{p}$ from $d\left(e\left(\tilde{x}\right)\right)$ and $d\left(e\left(\hat{x}\right)\right)$ according to $s_p$, $n_p$, $\tilde{y}$ and $\hat{y}$\\
	    \STATE Compute global style contrastive loss $\mathcal{L}_G$ using Equation~\ref{Global_loss}
	    \STATE Compute local matching contrastive loss $\mathcal{L}_L$ using  Equation~\ref{local_loss}
	    \STATE Compute total loss: $\mathcal{L}=\lambda \cdot \mathcal{L}_G+(1-\lambda)\mathcal{L}_L$
	    \STATE Update network $e(\cdot)$, $d(\cdot)$, $g(\cdot)$ and $g_L(\cdot)$ to minimize $\mathcal{L}$
	\ENDFOR
	\RETURN encoder $e(\cdot)$ and decoder $d(\cdot)$
	\end{algorithmic}
\end{algorithm}
\par
\begin{table*}[!htbp]
  \centering
  \footnotesize
  \caption{Description of the four datasets used in our experiment}
    \begin{tabular}{p{14em}cccc}
    \toprule
    \textbf{Datasets} & \multicolumn{1}{c}{Potsdam} & \multicolumn{1}{c}{DGLC} & \multicolumn{1}{c}{Hubei} & \multicolumn{1}{c}{Xiangtan} \\
    \midrule
    Ground resolution & 0.05m & 0.5m & 2m & 2m \\
    Spectral bands & NIR,RGB & RGB & RGB & RGB \\
    Crop size & $256\times256$ & $512\times512$ & $256\times256$ & $256\times 256$ \\
    Amount of self-supervised data & 13824 & 18248 & 66471 & 16051 \\
    Default amount of training data during fine-tuning(1\%) &  \multirow{2}[0]{*}{138}   & \multirow{2}[0]{*}{182}    & \multirow{2}[0]{*}{664}   & \multirow{2}[0]{*}{160} \\
    Amount of testing data & 1500  & 1825  & 9211  & 3815 \\
    \bottomrule
    \end{tabular}%
    \label{tab:dataset}%
\end{table*}%
Furthermore, we provide Algorithm~\ref{alg:GLCNet} to describe our proposed GLCNet in detail.

\section{Experiments and results}
\subsection{Data Description}
We evaluate the proposed GLCNet and other self-supervised methods on four datasets for RSI semantic segmentation. The ISPRS Potsdam Dataset and the Deep Globe Land Cover Classification Dataset are publicly available datasets. The Hubei and Xiangtan datasets come from the real world and have the same spatial resolution and a similar classification system that is convenient for studying the impact of domain differences. The details of the four datasets are explained below.
\subsubsection{ISPRS Potsdam Dataset}
The ISPRS Potsdam dataset consists of 38 high-resolution remote sensing aerial images. The images have a spatial resolution of 5 cm and 4 spectral bands: red, blue, green, and NIR. The dataset is annotated with six classes: low vegetation, trees, buildings, impervious surfaces, cars, and others. There are 38 patches with a size of 6000$\times$6000 pixels in the Potsdam dataset. Twenty-four patches cropped into 13824 images with a size of 256$\times$256 pixels are used for self-supervised training. To evaluate the performance of self-supervised learning, 1\% of the self-supervised training set is selected by default as the training set for the downstream semantic segmentation task, and the test set contains 1500 samples randomly selected from the remaining 14 patches that are cropped into 256$\times$256 pixels.
\subsubsection{Deep Globe Land Cover Classification Dataset (DGLC)}
The DeepGlobe Land Cover Classification dataset \cite{DemirKoperski-54-DGLC} provides high-resolution sub-meter satellite images with a size of 2448$\times$2448 pixels. The labels are far from perfect, covering seven classes: urban, agriculture, rangeland, forest, water, barren, and unknown. We select 730 images for training and 73 images for downstream testing. Moreover, each image is cropped to a size of 512$\times$512 pixels, and the final number of samples used for each stage is shown in Table \ref{tab:dataset}. 
\subsubsection{Hubei Dataset}
The images of the Hubei dataset are acquired from the Gaofen-2 satellite, covering the Hubei Province of China. The images have a spatial resolution of 2 m with three spectral bands(RGB). The labels are of poor quality, covering 10 classes: background, farmland, urban, rural areas, water, woodland, grassland, other artificial facilities, road, and others. We first divided the entire Hubei province into several patches with a size of 13889$\times$9259 pixels. From these patches, we randomly select 34 for training and 5 for testing. Moreover, each image is cropped to a size of 256$\times$256 pixels, and the final number of samples used for each stage is shown in Table \ref{tab:dataset}.
\subsubsection{Xiangtan Dataset}
The images of the Xiangtan dataset also come from the Gaofen-2 satellite and cover the city of Xiangtan, China. The labels are of higher quality than the Hubei dataset, covering 9 classes: background, farmland, urban, rural areas, water, woodland, grassland, road, and others. The entire city of Xiangtan is divided into patches with a size of 4096$\times$4096 pixels. We randomly select 85 of the patches for training and 21 for testing. Moreover, each image is cropped to a size of 256$\times$256 pixels, and the final number of samples used for each stage is shown in Table \ref{tab:dataset}.

\begin{figure*}[!ht]
    \vspace{0.15cm} 
    \setlength{\abovecaptionskip}{0.1cm} 
    \setlength{\belowcaptionskip}{0.15cm} 
    \centering
    \includegraphics[width=5in]{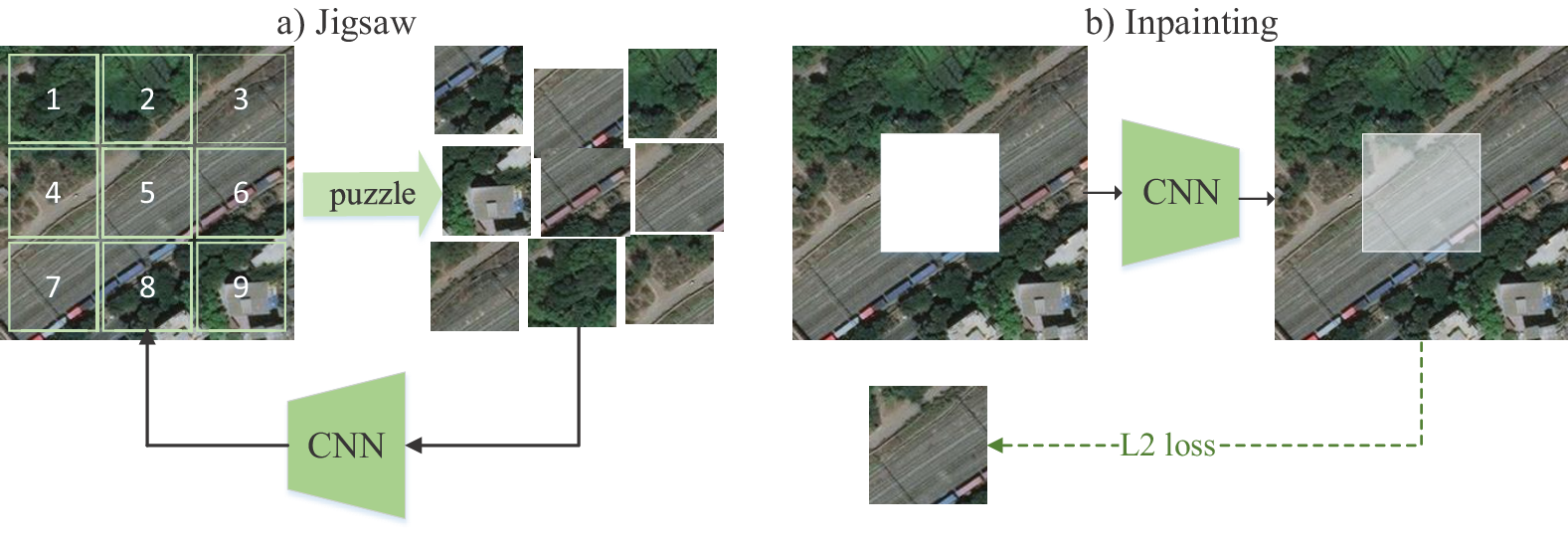}
    \caption{Diagram of two types of self-supervised pre-training tasks. a) Jigsaw, b) Image Inpainting}

	\label{fig:jig_inp}
\end{figure*}
\subsection{Experiment Setup}
\subsubsection{Baseline}
To evaluate the performance of self-supervised learning, we randomly initialize the network on specific downstream semantic segmentation tasks as the basic baseline. In addition, the common ImageNet pre-training strategy is also used as a baseline. Moreover, we compare 3 typical self-supervised tasks (predicting context \cite{DoerschGupta-65-predicting-context,NorooziFavaro-66-jigsaw}, image inpainting \cite{PathakKrahenbuhl-67-inpainting}, and instance-wise contrastive learning \cite{WuXiong-52-instance-discrimination,Chen2020-simCLR,Chen2020-moco}. The specific methods used for comparison are summarized as follows:
\begin{itemize}
\item[1.]
\textit{Random baseline:} Train from scratch on a specific semantic segmentation task without pre-training;
\item[2.]
\textit{ImageNet Pre-training:} Use the supervised training model on ImageNet to initialize the backbone of the semantic segmentation model;
\item[3.]
\textit{Jigsaw \cite{NorooziFavaro-66-jigsaw}:} This self-supervised method constructs self-supervised tasks by solving puzzles. Specifically, as shown in Fig.~\ref{fig:jig_inp} a), the given image is divided into multiple patches, and then the patch is shuffled before being sent to the CNN network. It is expected that the network will learn contextual relationships between the shuffled patches;
\item[4.]
\textit{Inpainting \cite{PathakKrahenbuhl-67-inpainting}:} A typical method of designing self-supervised signals with the idea of image restoration. Specifically, as shown in Fig.~\ref{fig:jig_inp} b), a random region of the image is first discarded, and then the CNN model is trained to predict the original image from the corrupted image, thus enabling the CNN model to learn contextual information;
\item[5.]
\textit{SimCLR\cite{Chen2020-simCLR}:} The SimCLR method is based on the idea of instance-wise contrastive learning, which learns by forcing positive samples enhanced from a sample to be similar and negative samples enhanced from different samples in a mini-batch to be dissimilar;
\item[6.]
\textit{MoCo v2 \cite{Chen2020-moco}:} MoCo v2 is also based on the idea of instance-wise contrastive learning, which focuses on obtaining negative samples far beyond the batch size. Therefore, a dynamic queue is proposed to save the features of negative samples, and a momentum update encoder is proposed to avoid the consistency problem of the representations of negative samples from the rapid change of the encoder.
\end{itemize}
\par

\subsubsection{Evaluation Metrics}
The performance of self-supervised methods needs to be evaluated on specific downstream semantic segmentation tasks. Therefore, we use OA and Kappa to measure the overall accuracy on the test dataset of downstream tasks, which are defined as follows:

\begin{equation}
OA=\frac{TP}{N}
\end{equation}
\begin{equation}
\text { Kappa }=\frac{OA-p_{e}}{1-p_{e}}
\end{equation}
where $TP$ denotes the total number of pixels that are correctly predicted, i.e., the true positives. $N$ represents the total number of pixels, $p_{e}=\frac{a_{1} \times b_{1}+\cdots+a_{c} \times b_{c}}{N \times N}$, $a_c$ denotes the actual number of pixels of class $c$, and $b_c$ denotes the number of predicted pixels of class $c$.
\par
In addition, we use the F1-score to measure the accuracy of a single category, which is defined as follows:

\begin{equation}
F_{1}=2 \times \frac{precision \times recall}{precision + recall}
\end{equation}
Where, $precision=\frac{TP}{TP+FP}$, $recall=\frac{TP}{TP+FN}$, $TP$ represents the true positives, $FP$ represents the false positives, and $FN$ represents the false negatives.
\subsubsection{Implementation Details}
\label{sec:details}
In the self-supervised pre-training phase, Jigsaw, SimCLR, and MoCo v2 are only designed to train the encoder of DeepLabV3+ with the ResNet50 backbone, while Inpainting and the proposed GLCNet train the complete encoder-decoder part of DeepLabV3+. We use the Adam optimizer for 400 epochs, with a batch size of 64. The initial learning rate is set as 0.01 with the cosine decay schedule. Moreover, for the proposed GLCNet method, we choose 4 local regions with a size of 48×48 from a sample, i.e., $s_p=48$, $n_p=4$. The model with the lowest loss in the self-supervised pre-training process is saved for the downstream task.
\par
Although Inpainting and the GLCNet method can train both encoder and decoder of the network during self-supervised training, methods such as SimCLR, which are used for comparison, are designed to train only the encoder. Therefore, we only load the encoder part from the self-supervised pre-training stage by default at the fine-tuning stage. In the fine-tuning phase, we use only a limited amount of annotated data for semantic segmentation training, such as 1\% of the amount of self-supervised data. We use the Adam optimizer for 150 epochs, with a batch size of 16. The initial learning rate is set as 0.001 and decreases to 0.98 every epoch.
\begin{table*}[ht]\footnotesize
  \centering
  \begin{threeparttable}
  \caption{Comparison with other methods using limited labeled data on four RSI semantic segmentation tasks}
  \begin{tabular}{lcccccccc}
    \toprule 
    \multirow{2}[4]{*}[0ex]{Pretext task} & \multicolumn{2}{c}{Potsdam} & \multicolumn{2}{c}{DGLC} & \multicolumn{2}{c}{Hubei} & \multicolumn{2}{c}{Xiangtan} \\
    \cmidrule{2-9}    
    \multicolumn{1}{r}{} & \multicolumn{1}{c}{Kappa} & \multicolumn{1}{c}{OA} & \multicolumn{1}{c}{Kappa} & \multicolumn{1}{c}{OA} & \multicolumn{1}{c}{Kappa} & \multicolumn{1}{c}{OA} & \multicolumn{1}{c}{Kappa} & \multicolumn{1}{c}{OA} \\
    \midrule
    Random\_Baseline & 58.27 & 67.39 & 51.47 & 71.70  & 46.69 & 58.02 & 64.85 & 78.17 \\
    ImageNet\_Supervised\_Baseline& 67.65 & 74.83 & 65.64 & 79.17 & \textbf{55.96} & \textbf{65.19} & 71.43 & 82.09 \\
    \midrule
    Jigsaw & 60.99 & 69.68 & 44.05 & 68.88 & 46.99 & 58.80  & 66.23 & 78.97 \\
    Inpainting & 63.62 & 71.70  & 43.82 & 67.78 & 44.47 & 56.80  & 71.15 & 81.89 \\
    MoCo v2 & 59.81 & 68.73 & 54.02 & 72.93 & 48.69 & 59.63 & 66.96 & 79.28 \\
    SimCLR & 65.62 & 73.21 & 67.33  & 79.77 & 51.85 & 62.12 & 70.76 & 81.55 \\
    \textbf{Ours(GLCNet)}  & \textbf{71.80} & \textbf{78.05} & \textbf{67.82} & \textbf{80.49} & \textbf{53.80} & \textbf{63.62} & \textbf{72.16} & \textbf{82.57} \\
    \bottomrule
    \end{tabular}%
    \label{tab:ex1}%
    \end{threeparttable}
\end{table*}%
\begin{figure*}[!ht]
    \vspace{0.15cm} 
    \setlength{\abovecaptionskip}{0.1cm} 
    \setlength{\belowcaptionskip}{0.15cm} 
    \centering
    \includegraphics[width=6in]{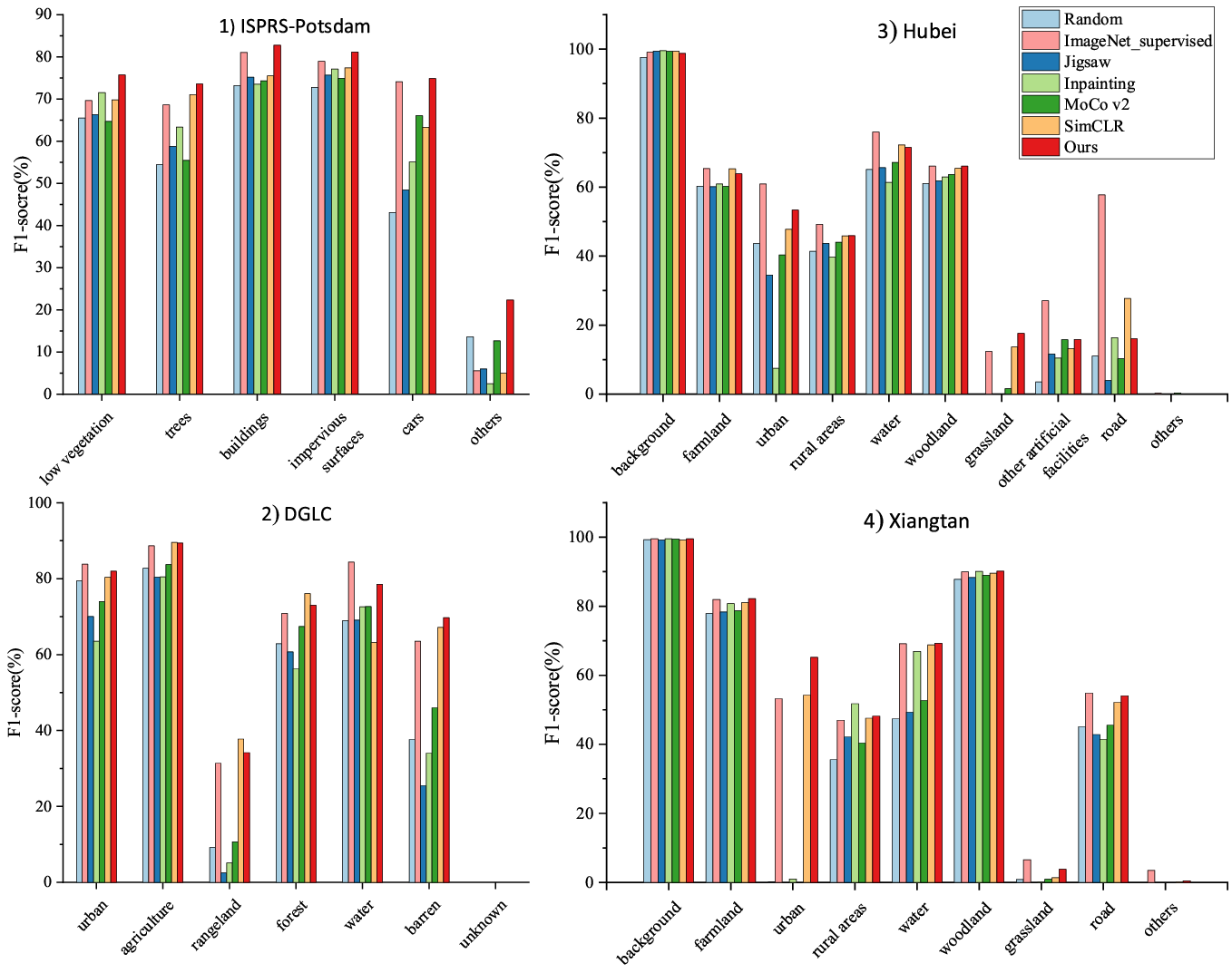}
    \caption{F1-score for each class on four RSI semantic segmentation tasks}
	\label{fig:single_class_result}
\end{figure*}
\subsection{Experimental Results}
In this section, we first compare the proposed GLCNet with other self-supervised methods and the ImageNet pre-training method on several RSI semantic segmentation datasets. Then, we explore two factors that may affect the self-supervised pre-training performance on the target RSI semantic segmentation task: the amount of self-supervised pre-training data and the domain differences between the pre-training dataset and the fine-tuning dataset.
\subsubsection{Comparison with other methods}

\begin{figure*}[htbp]
  
    \centering
    \includegraphics[width=4.5in]{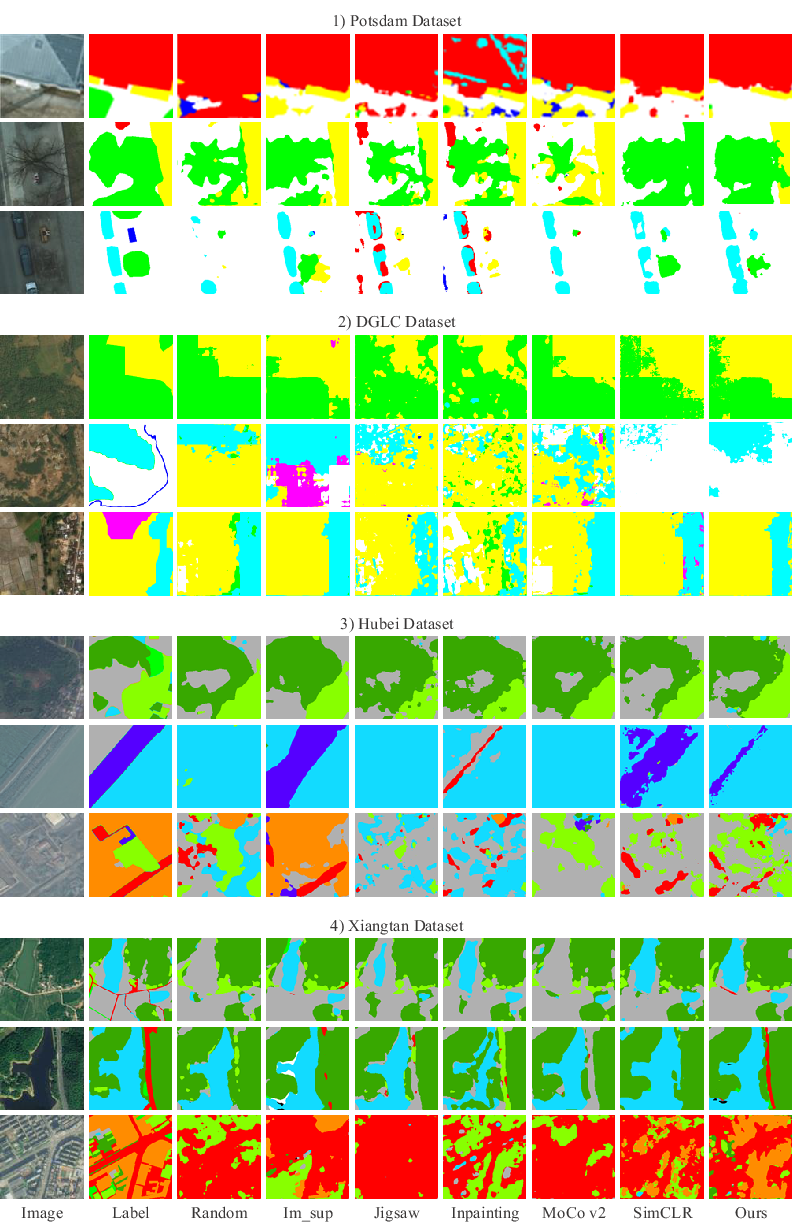}
    \caption{Examples of visualization results on four datasets. ( Im\_sup is ImageNet supervised pre-training )}
	\label{fig:vis_result}
\end{figure*}

In this section, we evaluate the performance of the proposed GLCNet on the RSI semantic segmentation task with limited annotations on multiple datasets and compare it with other self-supervised learning methods, ImageNet pre-training method, and random initialization method. The amount of data used for self-supervised pre-training on each dataset is shown in Table \ref{tab:dataset}, and 1\% of the amount of self-supervised data is used for fine-tuning. From the results in Table \ref{tab:ex1}, we find that our proposed GLCNet method greatly improves all datasets when compared to not implementing any pre-training strategy. At the same time, we find that the performance of different self-supervised methods is quite different, and inappropriate self-supervised methods have a negative impact, while our method achieves state-of-the-art results. As illustrated in Fig. \ref{fig:vis_result}, we also show some visualization results, where our method's performance is relatively better overall. Meanwhile, to measure whether our method has advantages in each category, we calculate the single-class accuracy, the results of which are shown in Fig. \ref{fig:single_class_result}. From Fig. \ref{fig:single_class_result}, we find that our method achieves superior results in most classes compared to other self-supervised methods on multiple datasets.
\par

In addition, our method outperforms the ImageNet pre-training method on most of the datasets, where the ImageNet pre-training method is obtained by supervised training on ImageNet with millions of data, which is much more than the amount of self-supervised data used in our experiments. This shows that although the ImageNet pre-training method can provide a significant improvement, it is not the optimal approach due to the large differences between natural images and remote sensing images. For example, as shown in Table \ref{tab:ex1}, our method on the Potsdam dataset has the most obvious improvement compared to ImageNet pre-training, possibly because the dataset has four bands, which is the most different from RGB natural images. 
Therefore, it is more reasonable to train a general model directly from unlabeled remote sensing images. Furthermore, it is worth noting that in our experiments, the images used for the self-supervised pre-training are similar to those used for the downstream task, and that both originate from the same dataset. Such a situation is available, as we can easily obtain a large number of images from the same source through satellite technology.
\subsubsection{Effect of the amount of self-supervised data}
\label{sec:self amount} 
\begin{figure}[ht]
  
    \centering
    \includegraphics[width=3.5in]{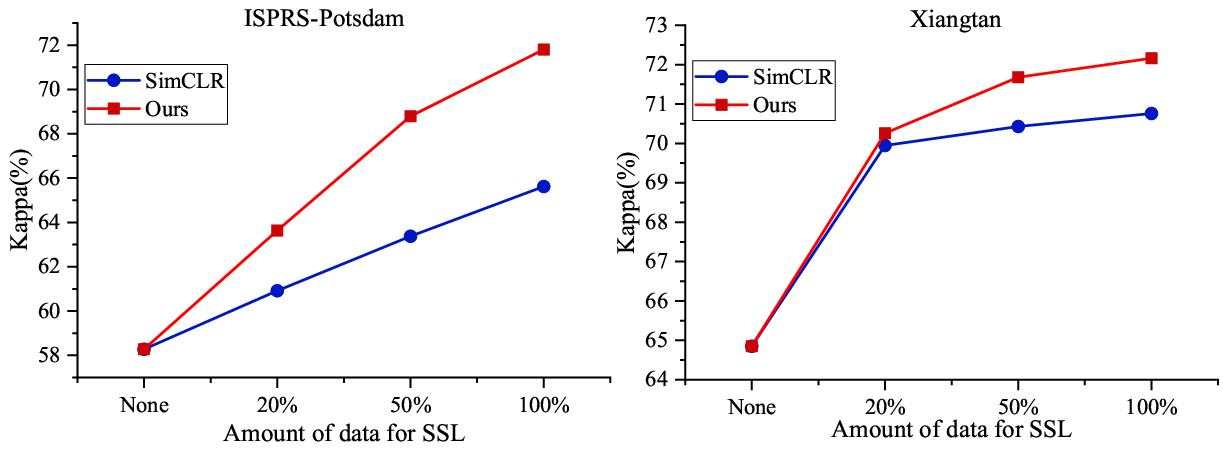}%
    \caption{Classification accuracies with different amounts of self-supervised data. 1) Results on the ISPRS-Potsdam dataset, 2) Results on the Xiangtan dataset}

	\label{fig:ssl_amount_result}
\end{figure}
Since self-supervised pre-training does not need annotated data and a large amount of image data is easy to obtain, this section mainly explores whether more self-supervised pre-training data can improve performance. To this end, we conduct experiments on the ISPRS-Potsdam dataset and the Xiangtan dataset by randomly selecting 20\%, 50\%, and 100\% of self-supervised data. The results are shown in Fig. \ref{fig:ssl_amount_result}, where None represents that no self-supervised training was performed. From the results, we find that in both datasets, there is an overall increasing trend as the amount of self-supervised data increases, and the improvement of our method is relatively more obvious compared with the SimCLR method. Therefore, it is foreseeable that the proposed method may be more beneficial when using larger datasets for self-supervised training.

\begin{table*}[ht]\footnotesize
  \centering
  \begin{threeparttable}
  \caption{Results on the domain difference between self-supervised pre-training dataset and fine-tuning dataset}
    \begin{tabular}{p{7.5em}<{\centering}lcccccc}
    \toprule
    \multirowcell{3}{Pre-training\\dataset} & \multicolumn{1}{c}{\multirow{3}[6]{*}{Method}} & \multicolumn{4}{c}{Fine-tuning dataset} & \multirowcell{3}{Domain\\ differences} \\
\cmidrule{3-6}          &       & \multicolumn{2}{c}{Xiangtan} & \multicolumn{2}{c}{Hubei} &  \\
   &       & Kappa & OA    & Kappa & OA    &  \\
    \midrule
    -     & Random\_Baseline & 64.85 & 78.17 & 46.69 & 58.02 & - \\
    \specialrule{0.05em}{1pt}{1pt}
    \multirow{3}[0]{*}{Postdam} & Supervised\_Baseline & 68.05 & 80.11 & 47.54 & 59.06 & \multirowcell{3}{Dissimilar\tnote{1}} \\
          & SimCLR & 66.80  & 79.24 & 47.69 & 58.91 &  \\
          & \textbf{GLCNet} & 68.37 & 80.18 & 48.94 & 59.95 &  \\
    \specialrule{0em}{1.5pt}{1pt}
    \multirow{3}[0]{*}{DGLC} & Supervised\_Baseline & 69.35 & 80.72 & 50.81 & 60.93 & \multirowcell{3}{Similar\tnote{2}} \\
          & SimCLR & 69.81 & 81.07 & 50.27 & 60.90 &  \\
          & \textbf{GLCNet} & 70.87 & 81.66 & 49.58 & 60.76 &  \\
    \specialrule{0em}{1.5pt}{1pt}
    \multirow{3}[0]{*}{Xiangtan} & Supervised\_Baseline & -     & -     & \textbf{53.53} & \textbf{63.63} & \multirowcell{3}{Highly\\similar\tnote{3}} \\
          & SimCLR & 70.76 & 81.55 & 49.20  & 60.20  &  \\
          & \textbf{GLCNet} & \textbf{72.16} & \textbf{82.57} & 51.16 & 61.94 &  \\
    \specialrule{0em}{1.5pt}{1pt}
    \multirow{3}[0]{*}{Hubei} & Supervised\_Baseline & 71.16 & 81.89 & -     & -     & \multirowcell{3}{Highly\\similar\tnote{3}} \\
          & SimCLR & 70.67 & 81.53 & 51.85 & 62.12 &  \\
          & \textbf{GLCNet} & 71.55 & 82.13 & \textbf{53.80} & \textbf{63.62} &  \\
    \specialrule{0em}{1.5pt}{1pt}
    \multirow{3}[0]{*}{Xiangtan+Hubei} & Supervised\_Baseline & -     & -  & - & - &
    \multirowcell{3}{Mixed\\domains} \\
    & SimCLR & 69.72 & 81.10  & 50.44 & 60.77 & \\
          & \textbf{GLCNet} & 70.81 & 81.81 & 53.11 & 63.00    &  \\
    \specialrule{0em}{1.5pt}{1pt}
    \multirowcell{3}{Xiangtan+Hubei\\+DGLC} & Supervised\_Baseline & -     & -  & - & - &
    \multirowcell{3}{Mixed\\domains} \\
    & SimCLR & 70.17 & 81.23 & 49.32 & 59.86 & \\
          & \textbf{GLCNet} & 71.19 & 81.76 & 51.62 & 62.05 &  \\
    \bottomrule
    \end{tabular}%
    \begin{tablenotes}   
        \footnotesize               
        \item[1] Different spectral bands; Large difference in image resolution; Far in physical location; Large difference in classification system       
        \item[2] Same spectral bands; Small difference in image resolution; Far in physical location; Some difference in classification system    
        \item[3] Same spectral bands; Same image resolution; Close in physical location; Similar classification system  
      \end{tablenotes}           
    \label{tab:domain}%
    \end{threeparttable}
\end{table*}%
\subsubsection{Effect of the domain difference}
\label{sec:domain effect}
In this section, we evaluate the impact of domain differences on the performance of the self-supervised pre-training model. The results are shown in Table \ref{tab:domain}, where '$Supervised\_Baseline$' indicates supervised training using the pre-training dataset first, and then transferring it to downstream tasks. From the results, we find that training with the self-supervised dataset, which is more similar to the downstream task dataset, led to better model performance on the downstream task. In addition, our method mostly outperforms supervised learning, except in cases where the domain differences are extremely small (e.g., Hubei$\rightarrow$Xiangtan, Xiangtan$\rightarrow$Hubei), mainly because the two domains not only have the same image resolution and are close in a physical location but also, crucially, have an approximately consistent classification system. Therefore, it is difficult to exceed the accuracy of supervised learning. Although we find in Section~\ref{sec:self amount} that model performance further improves as the amount of self-supervised training data increases, in this experiment we find that the model performance may not improve or may even be damaged if the self-supervised pre-training dataset is mixed with a large number of images that are not very similar to the downstream task dataset. Fortunately, since self-supervised pre-training does not require labels, it is feasible to obtain a large amount of image data that is similar to the target dataset.

\subsection{Ablation Study}\label{sec:ablation}
In this section, we perform ablation experiments to investigate the effectiveness of the modules of our proposed GLCNet method, the effectiveness of the decoder parameters of the models trained with our method, and the effect of different values of the loss weight $\lambda$.
\subsubsection{The effectiveness of the modules of the proposed GLCNet} 
\par
\begin{figure}[ht]
    \centering
    \includegraphics[width=3in]{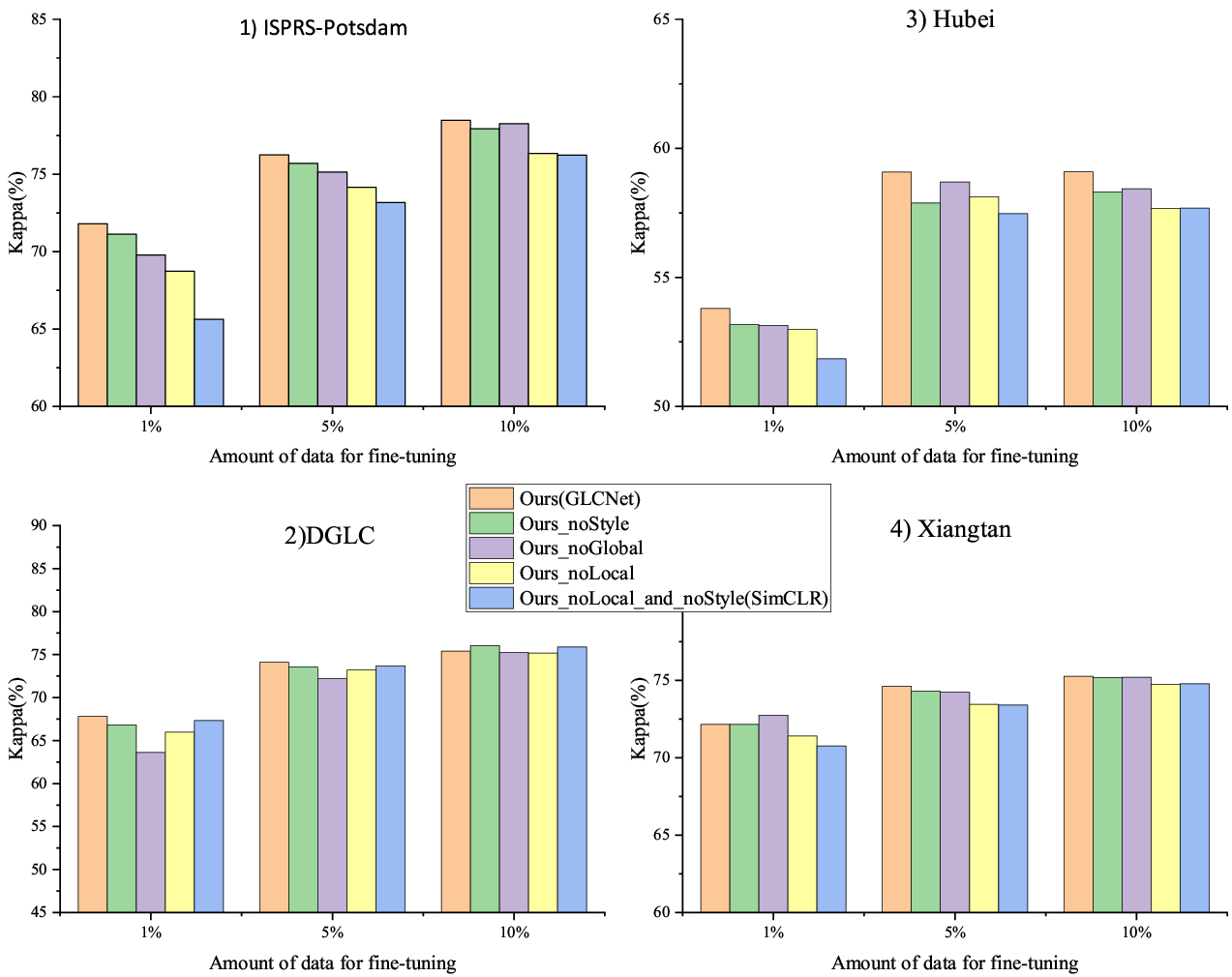}
    \caption{ Results of an ablation experiment exploring the effectiveness of the modules of the GLCNet}

	\label{fig:ablation_result}
\end{figure}

In this section, we explore the effectiveness of each module in our proposed method on four datasets. The experimental results are shown in Fig. \ref{fig:ablation_result}, where:
\par
$\bullet$ $Ours\_noStyle$ indicates the global module without using style features, i.e., directly using the traditional global average pooling features; 
\par
$\bullet$ $Ours\_noGlobal$ indicates that the global style contrastive learning module is completely removed;
\par
$\bullet$ $Ours\_noLocal$ indicates that the local matching contrastive learning module is completely removed;
\par
$\bullet$ $Ours\_noStyle\_and\_noLocal$ indicates that the local matching contrastive learning module is removed and that the global module does not use style features. 
\par

From the results, we find that the complete GLCNet achieves the optimal performance and each module has some benefits in most of the experiments. Moreover, those methods with a local matching contrastive learning module have significantly higher performance than those without on most datasets. Hence, local differentiation is necessary for practical RSI semantic segmentation tasks. However, surprisingly, $Ours\_noGlobal$ achieves the worst results on the DGLC dataset, which indicates that the global module is extremely important on this dataset. This may be due to the fact that local regions are overly distinguished when the method with only local matching contrastive learning module, while about half of the images in the DGLC dataset have only one category on a single image, so the method with only local contrastive learning module is particularly inappropriate in such a case.
\par
\begin{table}[!ht]\footnotesize
  \centering
  \caption{Results of loading different levels of GLCNet pre-training parameters during fine-tuning}
  \begin{tabular}{p{8.5em}cccc}
  \toprule
  \multirowcell{2}{Load pre-training \\parameters} & \multicolumn{2}{c}{Potsdam} & \multicolumn{2}{c}{Xiangtan} \\
  \cmidrule{2-5}    \multicolumn{1}{c}{} & \multicolumn{1}{c}{Kappa} & \multicolumn{1}{c}{OA} & \multicolumn{1}{c}{Kappa} & \multicolumn{1}{c}{OA} \\
  \midrule
  encoder & 71.80  & 78.05 & 72.16 & \textbf{82.57} \\
  encoder+d(1,2) & 72.02 & 78.13 & \textbf{72.25} & 82.52 \\
  encoder+d(1,2,3) & \textbf{72.17} & \textbf{78.29} & 71.54 & 82.19 \\
  \bottomrule
  \end{tabular}%
  \label{tab:ex-decoder}%
\end{table}%
\subsubsection{The effectiveness of the decoder part trained with GLCNet}
\par
Our method was originally designed to train the full semantic segmentation network, but since most of the methods used for comparison are designed to train only the encoder, we only loaded the self-supervised pre-trained encoder in the previous experiments during the fine-tuning phase for fairness of comparison. In this section, to investigate whether the decoder trained by our method is valid, we conduct experiments on the Potsdam and Xiangtan datasets. The experimental results are shown in Table \ref{tab:ex-decoder} , where $d(1,2)$ denotes loading the decoder parameters of the first two layers and $d(1,2,3)$ denotes loading the complete decoder parameters except for the final classification layer. From the results, we find that the decoder parameters trained with our method do not bring out significant improvement, which may be because the decoder of the semantic segmentation network is mainly used for detail recovery, while our current local matching contrastive learning module performs an average pooling operation on local regions to extract features, losing detailed information such as edge localization.

\subsubsection{Ablation study of weight $\lambda$}
\begin{figure}[ht]

    \centering
    \includegraphics[width=3in]{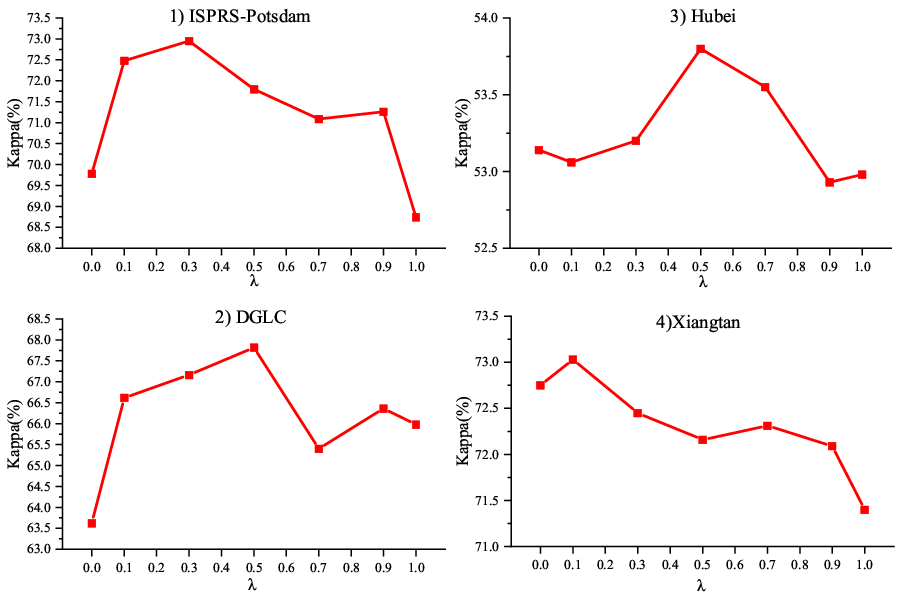}
    \caption{ Ablation study of weight $\lambda$. $\lambda =0$ means that there is only a local matching contrastive learning module during self-supervised training, and $\lambda =1$ means that there is only a global style contrastive learning module.}

	\label{fig:ablation_lamda}
\end{figure}

The hyper-parameter $\lambda$ in Equation \ref{total_loss} is set to balance the global style contrastive loss and the local matching contrastive loss. We believe that the global style contrastive learning module and the local matching contrastive learning module can learn different information useful for semantic segmentation tasks, both of which are very important, so we set $\lambda=0.5$ by default. However, this may not be optimal, so we further explore the effect of different $\lambda$ in this section. The experimental results are shown in Fig. \ref{fig:ablation_lamda}. From the experimental results, the optimal $\lambda$ is different on different datasets. In addition, the performance will drop rapidly in most datasets when our GLCNet method has only a local matching contrastive learning module, i.e. $\lambda$ = 0, or when our GLCNet method has only a global style contrastive learning module, i.e. $\lambda$ = 1. Therefore the final GLCNet should retain both the global style contrastive learning module and the local matching contrastive learning module. Overall, good performance is achieved on all datasets when $\lambda$ = 0.5.

\section{Discussions}
In this work, we apply self-supervised mechanisms to RSI semantic segmentation datasets, bringing significant improvements to RSI semantic segmentation tasks with insufficient annotated samples. We further discuss our experimental results in this section.
\par
We find that the design of the self-supervised task has a great impact on the final performance, and our proposed method achieves optimal results. Furthermore, through the experiments in Section \ref{sec:self amount}, we find that when the amount of self-supervised data increases, the fine-tuning accuracy is further improved. Therefore, it can be concluded that the model performance is expected to be further improved when more images are used for self-supervision; as a large number of remote sensing images are extremely easy to obtain, this will have great practical application value.
\par
The self-supervised trained model shows potential for remote sensing image understanding since it only depends on intrinsic supervisor signals instead of task-dependent labels. As the experimental results show in Section~\ref{sec:domain effect}, our proposed self-supervised approach outperforms supervised learning when there is some difference between the self-supervised and fine-tuned datasets, which illustrates that the model trained by self-supervised learning is more robust. In practice, we will face many situations where labels are lacking in some local areas. It will be extremely meaningful to learn a general model from images of the global area through self-supervised learning and then migrate to a local area. However, from the experimental results, we found that if multiple datasets with large differences are mixed for self-supervised training, the performance will be impaired when migrating to a local dataset. This may be because it is more difficult to perform self-supervised training when mixing multiple domains, as it tends to distinguish between images in different domains with large differences first, while the ability to distinguish between images in a local domain may be reduced. It would be of great practical value if we could subsequently find a way to mix images from multiple domains for self-supervised training without degrading its effectiveness in a single domain, so that we could perform self-supervised training to obtain a more general model by constructing a large image dataset composed of different resolutions, different regions, different times, etc.
\par
We find that each of the modules we designed has some benefit, while the local matching contrastive learning module brought huge improvements in most experiments, which illustrates the significance of local scale discrimination on the semantic segmentation dataset. However, $s_p$ and $n_p$ are the same for all datasets without much exploration, which may not be the optimal setting. Furthermore, as the distribution of surface features in the actual images can be extremely heterogeneous, the random selection of local areas will be biased toward more dominant feature classes. Further improvement might be achieved if the selection of local areas could be made more homogeneous.

\section{Conclusion}

In this work, we introduce self-supervised contrastive learning to RSI semantic segmentation tasks for learning general spatio-temporal invariant features from a large number of unlabeled images to reduce the dependence on labeled samples. Furthermore, considering that the existing contrastive learning methods are mainly designed for image classification tasks to obtain image-level representations, which may not be optimal for semantic segmentation tasks that require pixel-level discrimination, we propose the GLCNet. Experiments show that our method mostly outperforms the traditional ImageNet pre-training method and other self-supervised methods in semantic segmentation tasks with limited labeled data. We also find that more self-supervised pre-training samples can bring performance improvements, and in actual situations, we can easily obtain a large amount of remote sensing data, so our method may have great practical application significance.

There are still some shortcomings in our method; for example, we would like to use GLCNet to better learn general temporal invariance features. However, at present, we only simulate temporal transformations by randomly enhancing the images in terms of color and texture due to the lack of multi-temporal image data. This cannot really imitate the complex transformations caused by seasons, imaging conditions, etc. Therefore, the true temporal features might not be learned sufficiently, which can subsequently be complemented by using real multi-temporal images. In the future, the proposed method will be further improved and then applied to large-scale image data to alleviate the critical lack of labeling in tasks such as global land cover. Another potential research topic is to use adversarial examples \cite{ChenXiao-921,ChenXu-931,LiHuang-941} to improve the robustness of the pre-training model.

\section{Acknowledges}
We appropriate anonymous reviewers for their valuable suggestions to make this paper better. This work was supported in part by the National Natural Science Foundation of China under Grant 41871364, Grant 41861048 and Grant 42171376 and by the Fundamental Research Funds for the Central Universities of Central South University (2021zzts0842). We appropriate the High Performance Computing Platform of Central South University to provide HPC resources.

\ifCLASSOPTIONcaptionsoff
  \newpage
\fi

\bibliographystyle{IEEEtran}
\bibliography{Ref.bib}

\end{document}